\documentclass{article}

\PassOptionsToPackage{numbers, compress}{natbib}
\usepackage[final]{tackling_climate_workshop_style}

\usepackage[utf8]{inputenc} %
\usepackage[T1]{fontenc}    %
\usepackage{hyperref}       %
\usepackage{url}            %
\usepackage{booktabs}       %
\usepackage{amsfonts}       %
\usepackage{nicefrac}       %
\usepackage{microtype}      %

\usepackage{xcolor}
\usepackage{amsmath,amssymb}
\usepackage{graphicx}
\usepackage{colortbl}
\usepackage{siunitx}
\usepackage{enumitem}

\hypersetup{
    colorlinks,
    linkcolor={red!50!black},
    citecolor={blue!50!black},
    urlcolor={blue!80!black}
}

\newcommand{\greyrule}{\arrayrulecolor{black!30}\midrule\arrayrulecolor{black}}
\setlist[itemize]{leftmargin=0.6cm}
\setlist[enumerate]{leftmargin=0.6cm}

\title{Scene-to-Patch Earth Observation: Multiple Instance Learning for Land Cover Classification}

\author{
  Joseph Early\thanks{University of Southampton, UK;
  \texttt{\{J.A.Early,C.Evers,sdr1\}@soton.ac.uk}} 
 \And
   Ying-Jung Deweese\thanks{Georgia Institute of Technology, USA;
   \texttt{yingjungcd@gmail.com}}
 \And
   Christine Evers\footnotemark[1]
 \And
   Sarvapali Ramchurn\footnotemark[1]
}

\begin{document}

\maketitle

\vspace{-0.5cm}

\begin{abstract}
Land cover classification (LCC), and monitoring how land use changes over time, is an important process in climate change mitigation and adaptation. Existing approaches that use machine learning with Earth observation data for LCC rely on fully-annotated and segmented datasets. Creating these datasets requires a large amount of effort, and a lack of suitable datasets has become an obstacle in scaling the use of LCC. In this study, we propose Scene-to-Patch models: an alternative LCC approach utilising Multiple Instance Learning (MIL) that requires only high-level scene labels. This enables much faster development of new datasets whilst still providing segmentation through patch-level predictions, ultimately increasing the accessibility of using LCC for different scenarios. On the DeepGlobe-LCC dataset, our approach outperforms non-MIL baselines on both scene- and patch-level prediction. This work provides the foundation for expanding the use of LCC in climate change mitigation methods for technology, government, and academia.
\end{abstract}

\vspace{-0.3cm}
\section{Introduction}
\vspace{-0.1cm}
Land use/land cover (LULC) change has been recognised as a major contributor to the rise of atmospheric carbon dioxide (CO2). Changes in LULC have significant affects on the carbon balance within ecosystem services that contribute to climate change mitigation \cite{friedlingstein2020global}. LULC remote sensing (RS) techniques are widely used for areas such as sustainable development, crop health/yield monitoring, deforestation, urban planning, and water availability \cite{demir2018deepglobe, hoeser2020objectb}. Due to the recent curation of large volumes of accessible data, machine learning is seeing increased use in RS for Earth Observation (EO). In Land Cover Classification (LCC), the objective of machine learning is to identify and segment regions in satellite images according to a set of classes, e.g., agricultural land, urban land, water, etc. Typically, this requires a segmented dataset \textemdash{} a collection of images that have already been manually annotated with the different class regions. Creating these labelled datasets is a costly process: time and care must be taken to accurately segment the regions. This had lead to bottlenecks and concerns about a lack of datasets for machine learning in EO \cite{ccai2022, hoeser2020objecta}.
In this work, we present an alternative approach that does not require segmentation datasets.  Our contributions are as follows:

\vspace{-0.25cm}
\begin{enumerate}
    \itemsep-0.02cm 
    \item We reframe LCC as a Multiple Instance Learning (MIL) regression problem, reducing the need for segmentation labels whilst also preserving high-resolution data during training and inference.
    \item We propose Scene-to-Patch MIL models that transform low-resolution scene labels to high-resolution patch predictions for LCC.
    \item We explore how different data and model configurations of our approach affect performance on the popular DeepGlobe LCC dataset \cite{demir2018deepglobe}.\footnotemark[1]
\end{enumerate}
\vspace{-0.25cm}

The rest of this paper is laid out as follows. Section \ref{sec:background} provides the background to LCC, and Section \ref{sec:methodology} details our MIL approach. Our experiments are presented in Section \ref{sec:experiments}, and Section \ref{sec:conclusion} concludes.

\footnotetext[1]{Source code is available at \url{https://github.com/JAEarly/MIL-Land-Cover-Classification}}

\vspace{-0.1cm}
\section{Background and Related Work}
\label{sec:background}
\vspace{-0.1cm}

In this section, we provide the necessary background for our work, detailing LCC and its role in climate change mitigation, existing approaches to LCC, and a brief overview of MIL.

\vspace{-0.25cm}
\paragraph{Land Cover Classification} \hspace{-0.2cm} Machine learning for EO can be used in monitoring and forecasting of socioeconomic, ecological, agricultural, and hydrological problems \cite{liu2017deep}. For example, in the domain of monitoring green house gas (GHG) emissions, satellite images can be used to track emission sources and carbon sequestration \cite{rolnick2022tackling}. It is estimated that human land use contributes about a quarter of global GHG emissions, and that a reduction of around a third of emissions could come from better land management \cite{rolnick2022tackling}. As such, better understanding about how land is used can contribute towards zero emission targets. Improved policy design, planning, and enforcement can be achieved through real-time monitoring \cite{kaack2022aligning}. For example, automated LCC can be used to determine the effect of regulation or incentives to drive better land use practices \cite{rolnick2022tackling}. LCC can also be used to detect the amount of acreage in use for farmland in order to assess food security and GHG emissions \cite{ullah2022quantifying}.

\vspace{-0.25cm}
\paragraph{Existing LCC Approaches} \hspace{-0.2cm} LCC is typically approached as an image segmentation problem. The objective is to perform pixel-wise classification of an input image, such that each pixel is assigned a label from a set of classes, in effect creating a new image where different regions in the original image have been separated and classified. This requires the original images to be segmented prior to training, i.e., all pixels must be annotated with a ground-truth class. Popular approaches include Fully Convolutional Networks \cite{long2015fully}, U-Net \cite{ronneberger2015u}, and Pyramid Networks \cite{lin2017feature}. Existing works have applied these or similar approaches to LCC \cite{karra2021global, kuo2018deep,  rakhlin2018land, seferbekov2018feature, tong2020land, wang2020weakly}. We refer readers to \cite{hoeser2020objecta} for a more in-depth review of existing work.

\vspace{-0.25cm}
\paragraph{Multiple Instance Learning} \hspace{-0.2cm} In MIL, data are grouped into bags of instances \cite{carbonneau2018multiple}. In comparison to conventional supervised learning, where each instance has a label, in MIL, only bag-level labels are provided. This reduces the burden of labelling, as only the bags need to be labelled, not every instance. MIL has seen some existing use with EO observation data, for example fusing panchromatic and multi-spectral images \cite{liu2017deep}, settlement extraction \cite{vatsavai2013complex}, landslide mapping \cite{zhang2020deep}, and crop yield prediction \cite{wang2011mixture}. However, to the best of the authors' knowledge, no prior work has studied the use of MIL for generic multi-class LCC. We discuss our approach to this problem in the next section.

\vspace{-0.2cm}
\section{Methodology}
\label{sec:methodology}
\vspace{-0.2cm}

In this section, we propose our MIL approach to LCC. We first explain the process and benefits of reframing LCC as a MIL regression problem, and then detail the particular approach that we use.

\vspace{-0.2cm}
\subsection{Multiple Instance Learning for Land Cover Classification} 
\vspace{-0.2cm}

For EO images, we identify three distinct tiers of data: \textit{scene level} at the scale of the original images, \textit{patch level} at the scale of small blocks of the original image (typically tens or hundreds of pixels), and \textit{pixel level} at the scale of individual pixels. Segmentation-based models operate at the pixel level, and as such require pixel-level annotations. We propose a MIL-based approach that operates at the scene and patch levels. Instead of using pixel-level annotations, our approach only requires scene-level labels, i.e., for each input image, we only provide the proportion of each land cover class in that image. With these scene-level labels, LCC now becomes a regression problem, with the task of predicting the coverage proportion of each class. The primary motivation for such an approach is that it enables fast acquisition of training labels, as cost- and time-intensive pixel annotations are not required. It also reduces the likelihood of label errors, which often occur in EO datasets \cite{rakhlin2018land}.

\begin{figure}[htb!]
    \centering
    \includegraphics[width=0.95\textwidth]{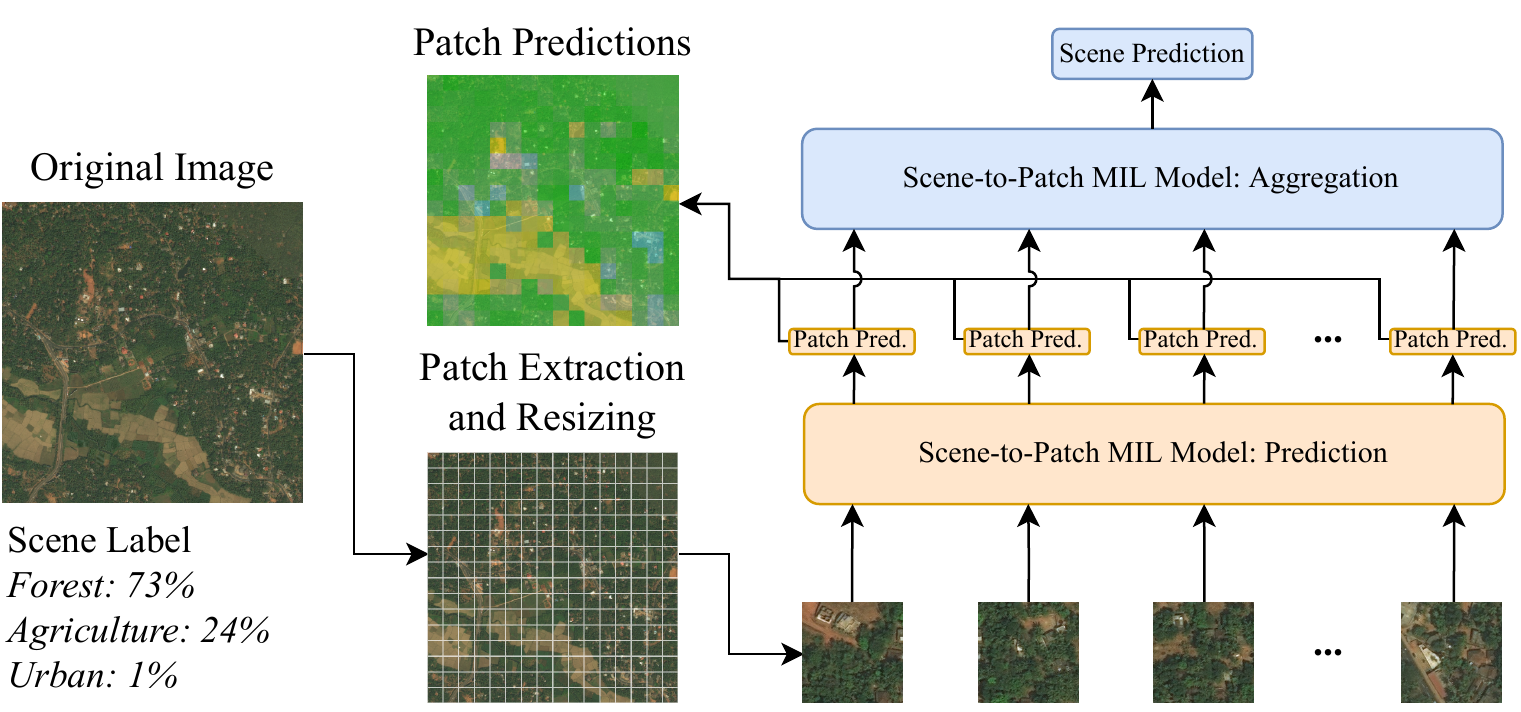}
    \vspace{-0.1cm}
    \caption{An overview of our MIL Scene-to-Patch approach for LCC. As the model produces both instance (patch) and bag (scene) predictions, scene-level labels are transformed to patch-level outputs.}
    \vspace{-0.2cm}
    \label{fig:overview}
\end{figure}

This reframed LCC problem does not necessitate a MIL approach \textemdash{} it can be treated as a purely supervised regression problem. However, as satellite images are often very high resolution, the images would have to be downsampled, and, as such, important data would be lost. With MIL, it is possible to operate at higher resolutions than a supervised learning approach. As depicted in Figure \ref{fig:overview}, our proposed MIL approach involves extracting patches from the original image using a grid, and then applying a MIL model to process and aggregate patch information. The MIL approach is operating on patch-level data while using scene-level labels, i.e., the images are converted into bags of patches, where each patch is an instance in MIL terms. 

\vspace{-0.2cm}
\subsection{Scene-to-Patch Models for Land Cover Classification}
\label{sec:model}
\vspace{-0.2cm}

Our proposed MIL approach utilises an Instance Space Neural Network (known as mi-Net in  \cite{wang2018revisiting}). We call these models Scene-to-Patch (S2P) classifiers as they learn from scene-level labels, but are able to generate both scene- and patch-level predictions, i.e., they upscale from low-resolution training labels to higher-resolution outputs. For a bag of patches extracted from a satellite image, an S2P model makes a prediction for each patch, and then takes the mean of these patch-level predictions to make an overall scene-level prediction. These S2P models can be relatively small (in comparison to baseline architectures such as ResNet18 and UNet; see Appendix \ref{app:model}), which means they are less expensive to train and run. This reduces the GHG emission effect of model development and use, which is an increasingly important consideration for the use of machine learning \cite{kaack2022aligning}. Other MIL models (e.g., with attention \cite{ilse2018attention}) could be used, but these would require post-hoc interpretability \cite{early2022model}.

\vspace{-0.3cm}
\section{Experiments}
\label{sec:experiments}
\vspace{-0.2cm}

In this section we give detail our experiments. First we introduce the dataset and model configurations used in this work (Section \ref{sec:dataset-and-models}), then provide our results and a discussion (Section \ref{sec:results}).

\vspace{-0.2cm}
\subsection{Dataset and Models}
\label{sec:dataset-and-models}
\vspace{-0.2cm}

Several datasets exist for using machine learning with EO \cite[p.~30]{hoeser2020objecta}. In this work we use the DeepGlobe-LCC dataset \cite{demir2018deepglobe} \textemdash{} an image segmentation LULC dataset with data sourced from the WorldView3 satellite (for more details, see Appendix \ref{app:data}). When transforming these EO images to MIL bags, there are two parameters to be determined: the size of the grid applied over the image (grid size), and the size that each cell of the grid is resized to (patch size). We experimented with three grid sizes (8, 16, and 24), and three patch sizes (small=28, medium=56, and large=102), resulting in nine different S2P configurations. Each patch size uses a slightly different model architecture. We compared our MIL S2P models to a fine-tuned ResNet18 model \cite{he2016deep}, and two UNet variations operating on different image sizes (224 x 224 and 448 x 448). These baseline models are trained in the same manner as the S2P models, i.e., using scene-level regression. Although the ResNet model does not produce patch- or pixel-level predictions, we use it as a scene-level baseline as many existing LCC approaches utilise ResNet architectures \cite{hoeser2020objecta, hoeser2020objectb}. For the UNet models, we follow the same procedure as \cite{wang2020weakly} and use class activation maps to recover segmentation outputs. This makes the UNet approach a stronger baseline than ResNet as it can be used for both scene- and pixel-level prediction. For more details on the models and training process, see Appendix \ref{app:model}.

\vspace{-0.2cm}
\subsection{Results}
\label{sec:results}
\vspace{-0.2cm}

\begin{figure}[htb!]
    \centering
    \includegraphics[width=\textwidth]{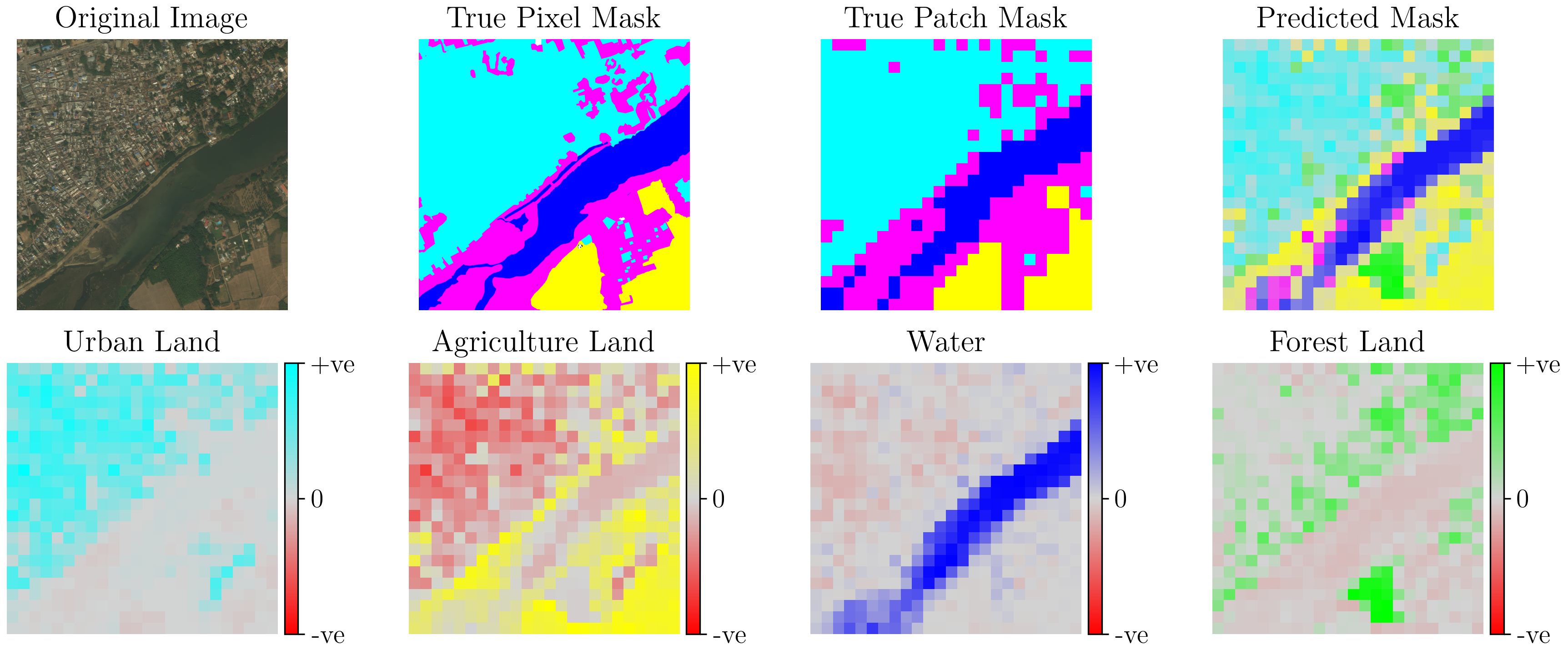}
    \vspace{-0.6cm}
    \caption{Example output of our S2P model (Medium 24). \textit{Top}: Original image, ground truth masks, and overall predicted mask. \textit{Bottom}: A per-class breakdown of the model predictions. Note how the model is able to both support (+ve) and refute (-ve) different classes for different regions of the image. There are also trees in the original image that have not been labelled as Forest land, but the model has been able to identify them. We provide further examples in Appendix \ref{sec:app_interpretability}.}  
    \label{fig:interpretability}
\end{figure}

We evaluate performance on scene-, patch-, and pixel-level prediction. For scene-level predictions, we report the root mean square error (RMSE) and mean average error (MAE), where lower values are better. For patch-level predictions, we report the patch-level mean Intersection over Union (mIoU; \cite{everingham2010pascal, minaee2021image}), where larger values are better. For pixel-level prediction, we compute pixel-level mIoU using the original ground truth segmentation masks (see Figure \ref{fig:interpretability}). The pixel segmentation metric is preferred over the patch metric as it is independent of grid size and compares to the highest resolution ground truth segmentation. Strong models should perform well at both scene- and pixel-level prediction, i.e., low scene RMSE and high pixel mIoU. Our results are given in Table \ref{tab:results}.

\vspace{-0.1cm}

{
\begin{table}[htb!]
    \centering
    \small
    \caption{Results for our nine MIL Scene-To-Patch (S2P) models and ResNet18 baseline. All results are given on the test dataset, and five repeats were conducted using 5-fold cross validation.}
    \vspace{-0.1cm}
    \label{tab:results}
    \begin{tabular}{lllll}
        \toprule
        Configuration & Scene RMSE & Scene MAE & Patch mIoU & Pixel mIoU \\
        \midrule
        ResNet18 & 0.218 $\pm$ 0.008 & 0.128 $\pm$ 0.004 & N/A & N/A \\
        UNet 224 & 0.136 $\pm$ 0.008 & 0.075 $\pm$ 0.004 & N/A & 0.245 $\pm$ 0.008 \\
        UNet 448 & 0.114 $\pm$ 0.006 & 0.064 $\pm$ 0.002 & N/A & 0.290 $\pm$ 0.010 \\
        \greyrule
        S2P Small 8 & 0.107 $\pm$ 0.003 & 0.058 $\pm$ 0.002 & 0.366 $\pm$ 0.015 & 0.337 $\pm$ 0.014 \\
        S2P Medium 8 & 0.098 $\pm$ 0.004 & 0.054 $\pm$ 0.002 & 0.414 $\pm$ 0.014 & 0.375 $\pm$ 0.013 \\
        S2P Large 8 & \textbf{0.090 $\pm$ 0.005} & \textbf{0.047 $\pm$ 0.002} & \textbf{0.439 $\pm$ 0.014} & \textbf{0.397 $\pm$ 0.014} \\
        \greyrule
        S2P Small 16 & 0.112 $\pm$ 0.007 & 0.059 $\pm$ 0.004 & 0.317 $\pm$ 0.007 & 0.305 $\pm$ 0.007 \\
        S2P Medium 16 & 0.093 $\pm$ 0.005 & 0.051 $\pm$ 0.003 & 0.407 $\pm$ 0.013 & 0.388 $\pm$ 0.012 \\
        S2P Large 16 & 0.097 $\pm$ 0.003 & 0.051 $\pm$ 0.001 & 0.404 $\pm$ 0.016 & 0.384 $\pm$ 0.014 \\
        \greyrule
        S2P Small 24 & 0.099 $\pm$ 0.004 & 0.052 $\pm$ 0.002 & 0.371 $\pm$ 0.012 & 0.360 $\pm$ 0.011 \\
        S2P Medium 24 & 0.098 $\pm$ 0.004 & 0.053 $\pm$ 0.002 & 0.379 $\pm$ 0.015 & 0.367 $\pm$ 0.015 \\
        S2P Large 24 & 0.106 $\pm$ 0.009 & 0.056 $\pm$ 0.005 & 0.353 $\pm$ 0.006 & 0.343 $\pm$ 0.006 \\
        \bottomrule
    \end{tabular}
    \vspace{-0.5cm}
\end{table}
}

From these results, we observe that the S2P Large model with a grid size of 8 is the most effective across all metrics. The ResNet18 model performs significantly worse than all of the S2P models on scene-level prediction. The S2P models all decrease in performance from patch to pixel mIoU, which is to be expected as they are unable to produce the fine details required in pixel-level segmentation. Despite this, they still outperform the UNet models. For the most part, smaller grid sizes lead to better performance. However, larger grid sizes give higher-resolution image segmentation outputs, so there is a trade-off. We suspect that smaller grid sizes are more effective as they are an inherent form of regularisation \textemdash{} larger grid sizes mean larger bags, which allow the model to more easily overfit. There is also the consideration that grid sizes that are too large may not capture sufficient information in each cell to facilitate accurate classification. We give an example of the outputs in Figure \ref{fig:interpretability}.

\vspace{-0.2cm}

\vspace{-0.45cm}
\section{Conclusions and Future Work}
\label{sec:conclusion}
\vspace{-0.2cm}

In this work, we proposed a new approach to LCC. Our method involves using MIL models as Scene-to-Patch classifiers that transform low-resolution training labels into high-resolution predictions. This overcomes the need for segmentation labels and will help accelerate the development of further EO datasets by reducing the need for costly and time-consuming labelling of EO data. Our approach can be improved in several ways: further optimising model architectures, using multiple grid sizes in a single model, and utilising spatial relationships to improve predictive performance. Ultimately, our work is a baseline exploratory approach. By enabling faster and easier curation of larger and more diverse datasets, it is hoped this foundational work will lead to further developments in the use of LCC for climate change mitigation in technology, government, and academia.

\begin{ack}
This work was funded by the AXA Research Fund and the UKRI Trustworthy Autonomous Systems Hub (EP/V00784X/1). We would like to thank the University of Southampton and the Alan Turing Institute for their support. The authors acknowledge the use of the IRIDIS high-performance computing facilities and associated support services at the University of Southampton in the completion of this work. Finally, this work was supported by the Climate Change AI (CCAI) Mentorship Program.
\end{ack}

\bibliographystyle{abbrvnat}
{\fontsize{9}{0}\bibliography{bib}}

\newpage
\appendix

\setcounter{table}{0}
\renewcommand{\thetable}{A\arabic{table}}

\setcounter{figure}{0}
\renewcommand{\thefigure}{A\arabic{figure}}

\section{Implementation and Resource Details}
\vspace{-0.2cm}

This work was implemented in Python 3.10 and the machine learning functionality used PyTorch. All the libraries used are detailed in the Github repository for this work, which can be found at \url{https://github.com/JAEarly/MIL-Land-Cover-Classification}. The majority of model training was carried out on a remote GPU service using a Volta V100 Enterprise Compute GPU with 16GB of VRAM, which utilised CUDA v11.0 to enable GPU support.
Training each model took a maximum of three and a half hours. Trained models can be found alongside the code in the Github repository. Fixed seeds were used to ensure consistency of dataset splits between training and testing; these are included in the scripts that are used to run the experiments. We used Weights and Biases \cite{wandb} to track our experiments, along with Optuna for hyperparameter optimisation \cite{akiba2019optuna}. During hyperparamater optimisation, we ran 40 trials with pruning using the Tree-structured Parzen Estimator sampler \cite{bergstra2011algorithms}.

\vspace{-0.2cm}
\section{Dataset}
\label{app:data}
\vspace{-0.2cm}

The DeepGlobe-LCC dataset is openly available and can be acquired from Kaggle. Below we give further details on the dataset, which are adapted from the Kaggle page.\footnote[2]{\url{https://www.kaggle.com/datasets/balraj98/deepglobe-land-cover-classification-dataset}}

\vspace{-0.2cm}
\subsection*{Data}
\vspace{-0.2cm}
The DeepGlobe-LCC dataset consists of 803 satellite images with 3 channels: red, green, and blue (RGB). Each image is 2448 x 2448 pixels with 50cm pixel resolution. All images were sourced from the WorldView3 satellite covering regions in Thailand, Indonesia, and India. The original challenge also had validation and test datasets with 171 and 172 images respectively, but as these datasets did not include masks, they were not used in this work.

\vspace{-0.2cm}
\subsection*{Labels}
\vspace{-0.2cm}
Each satellite image is paired with a mask image for land cover annotation. Each mask is an image with 7 classes of labels, using colour-coding (RGB) as follows:
\vspace{-0.2cm}

\begin{itemize}
\item \textbf{Urban land} (0, 255, 255) \textemdash{} Man-made, built up areas with human artefacts (ignoring roads which are hard to label).
\item \textbf{Agriculture land} (255, 255, 0) \textemdash{} Farms, any planned (i.e., regular) plantation, cropland, orchards, vineyards, nurseries, and ornamental horticultural areas.
\item \textbf{Rangeland} (255, 0, 255) \textemdash{} Any non-forest, non-farm, green land, grass.
\item \textbf{Forest land} (0, 255, 0) \textemdash{} Any land with x\% tree crown density plus clearcuts.
\item \textbf{Water} (0, 0, 255) \textemdash{} Rivers, oceans, lakes, wetland, ponds.
\item \textbf{Barren land} (255, 255, 255) \textemdash{} Mountain, land, rock, dessert, beach, no vegetation.
\item \textbf{Unknown} (0, 0, 0) \textemdash{} Clouds and others.
\end{itemize}

\vspace{-0.2cm}
\subsection*{Terms and Conditions}
\vspace{-0.2cm}
The DeepGlobe Land Cover Classification Challenge and dataset are governed by DeepGlobe Rules, The DigitalGlobe's Internal Use License Agreement, and Annotation License Agreement.

\vspace{-0.2cm}
\subsection*{Further Details}
\vspace{-0.2cm}

While the DeepGlobe-LCC dataset provides pixel-level annotations, these segmentation labels are \textit{only} used to generate the regression targets for our training and for the evaluation of derived patch segmentation, i.e., they are not used during training. However, we would like to stress that these segmentation labels are not strictly required for our approach, i.e., the scene-level regression targets can be created without having to perform segmentation.

We used 5-fold cross validation rather than the standard 10-fold due to the limited size of the datasets (only 803 images). With this configuration, each fold had an 80/10/10 split for train/validation/test. We normalised the images by the dataset mean (0.4082, 0.3791, 0.2816) and standard deviation (0.06722, 0.04668, 0.04768). No other data augmentation was used.

\section{Models and Training}
\label{app:model}
\vspace{-0.2cm}

In this section we give a formal definition of the MIL S2P models (\ref{app:formal_model}), then detail the model configurations (\ref{app:model_config}), architectures (\ref{app:s2p_model_arch}), and training procedures (\ref{app:training}).

\vspace{-0.2cm}
\subsection{Formal Model Definition}
\label{app:formal_model}
\vspace{-0.1cm}

For a collection of $n$ satellite images $\mathcal{X} = \{X_1,\ldots,X_n\}$, each image $X_i \in \mathcal{X}$ has corresponding label $Y_i \in \mathcal{Y}$, where $Y_i = \{Y_i^1,\ldots,Y_i^C\}$. $C$ is the number of land cover classes, $Y_i^c$ is the coverage proportion for class $c$ in image $X_i$, and $\sum_{c=1}^C Y_i^c = 1$. For an input image $X_i = \{x_i^1,\ldots,x_i^k\}$, where $x_i^j \in X_i$ is a patch extracted from the original image, our proposed Scene-to-Patch models make a prediction $\hat{y}_i^j$ for each patch, and then take the mean of these patch-level predictions to make an overall scene-level prediction $\hat{Y}_i = \frac{1}{k} \sum_{j=1}^k \hat{y}_i^j$. Note these models are trained entirely end-to-end and only use scene-level labels \textemdash{} no patch-level labels are used during training.

\vspace{-0.2cm}
\subsection{Model Configurations}
\label{app:model_config}
\vspace{-0.1cm}

We use twelve different model configurations in this work: one ResNet18 fully supervised approach, two UNet models, and nine different configurations of our Scene-to-Patch (S2P) approach. A summary of these configurations is given in Table \ref{tab:config}, and further details are given below.

\vspace{-0.2cm}
{
    \setlength{\tabcolsep}{5pt}
    \begin{table}[htb!]
    	\centering
    	\small
    	\caption{The ten different configurations used in this work. The grid size determines the number of cells and the size of each cell. Each cell is then resized (patch size), leading to a reduction in the overall image size (effective resolution and scale). \# Params is the number of parameters in each model; see \ref{app:s2p_model_arch} for more details regarding the \# Params in our S2P models.}
        \label{tab:config}
    	\begin{tabular}{lllllll}
            \toprule
			Configuration & Grid Size & Cell Size & Patch Size & Eff. Resolution & Scale & \# Params \\
			\midrule
			ResNet18 & 1 x 1 & 2448 x 2448 px & 224 x 224 px & 224 x 224 px & 0.8\% & 11.2M \\
                UNet 224 & 1 x 1 & 2448 x 2448 px & 224 x 224 px & 224 x 224 px & 0.8\% & 4.3M \\
			UNet 448 & 1 x 1 & 2448 x 2448 px & 448 x 448 px & 448 x 448 px & 3.3\% & 7.8M \\
            \greyrule
			S2P Small 8 & 8 x 8 & 306 x 306 px & 28 x 28 px & 224 x 224 px & 0.8\% & 707K \\
			S2P Medium 8 & 8 x 8 & 306 x 306 px & 56 x 56 px & 448 x 448 px & 3.3\% & 3.6M \\
			S2P Large 8 & 8 x 8 & 306 x 306 px & 102 x 102 px & 816 x 816 px & 11.1\% & 3.0M \\
           \greyrule
			S2P Small 16 & 16 x 16 & 153 x 153 px & 28 x 28 px & 448 x 448 px & 3.3\% & 707K \\
			S2P Medium 16 & 16 x 16 & 153 x 153 px & 56 x 56 px & 896 x 896 px & 13.4\% & 3.6M \\
			S2P Large 16 & 16 x 16 & 153 x 153 px & 102 x 102 px & 1632 x 1632 px & 44.4\% & 3.0M \\
           \greyrule
			S2P Small 24 & 24 x 24 & 102 x 102 px & 28 x 28 px & 672 x 672 px & 7.5\% & 707K \\
			S2P Medium 24 & 24 x 24 & 102 x 102 px & 56 x 56 px & 1344 x 1344 px & 30.1\% & 3.6M \\
			S2P Large 24 & 24 x 24 & 102 x 102 px & 102 x 102 px & 2448 x 2448 px & 100.0\% & 3.0M \\
			\bottomrule
        \end{tabular}
    \end{table}
}

\vspace{-0.5cm}
\paragraph{ResNet18} \hspace{-0.2cm} For the ResNet18 model, we treat our LCC regression problem in a fully supervised manner, i.e., without using a MIL approach. Instead, the entire satellite image is resized to 224 x 224 px (the size that ResNet18 expects), and the model makes (only) a scene-level prediction. Conceptually, this is equivalent to using a grid size of one and a patch size of 224 x 224 px (see Table \ref{tab:config}). We used a pre-trained ResNet18 model, with weights sourced from TorchVision.\footnote[3]{\url{https://pytorch.org/vision/stable/models/generated/torchvision.models.resnet18.html\#torchvision.models.resnet18}} We replaced the final classifier layer of the network with a new linear layer of seven outputs, and then re-trained the entire network, i.e., no weights were frozen during re-training.

\vspace{-0.3cm}
\paragraph{UNet Models} \hspace{-0.2cm} We used two different UNet configurations \textemdash{} one using entire image inputs resized to 224 x 224 px, and the other 448 x 448 px. The model makes scene-level predictions using global average pooling over $F$ (the output of the UNet's final convolutional layer) followed by a single classification layer $L$. Using class activation maps, it is possible to recover pixel-level segmentation outputs: $M_c = W_c F + B_c$, where $M_c$ is the class activation map for class $c$, $W_c$ are the weights in $L$ for class $c$, and $B_c$ is the bias for class $c$ in $L$. Note, for the UNet upsampling process, we experimented with fixed bilinear or learnt convolutions upsampling; the latter increases the number of model parameters. This was included as a hyperparameter during tuning, and it was found that fixed upsampling was best for the UNet 224 architecture, but learnt upsampling was best for UNet 448 architecture, leading to an increase in the number of parameters for the UNet 448 model.

\vspace{-0.2cm}
\paragraph{S2P Models} \hspace{-0.2cm} We tested nine different configurations of our S2P models. Two parameters were changed: the grid size (8, 16, or 24), and the patch size (small=28, medium=56, or large=102). The grid size determines the number of extracted patches. The patch size determines the model architecture that was used, i.e., we designed three different architectures, one for each patch size. This means models with different grid sizes but the same patch size used the same architecture, e.g., the S2P Large 8, S2P Large 16, and S2P Large 24 models all used the same model architecture (hence having the same number of model parameters in Table \ref{tab:config}).

\subsection{S2P Model Architectures}
\label{app:s2p_model_arch}
\vspace{-0.2cm}

The S2P models all used a consistent architecture: a feature extractor (convolutional and pooling layers), followed by a patch classifier (fully connected layers), and finally a MIL mean aggregator. The output of the classifier is a 7-dimensional vector, which represents the prediction for each class. Each patch is passed independently through the feature extractor + patch classifier to produce a prediction for each class, and then MIL mean aggregation is used to produce a scene-level prediction. The different architectures are given in Tables \ref{tab:s2p_small_arch} - \ref{tab:s2p_large_arch}. Note that, despite using larger patches, the S2P large architecture has fewer parameters than the S2P Medium architecture (see Table \ref{tab:config}) as it has an additional convolutional and pooling layer, leading to a smaller embedding size (5600 rather than 6912). 

\vfill

\vspace{-0.2cm}
\begin{table}[!htb]
	\centering
    \caption{S2P Small Architecture; patch size 28. For the Conv2d and MaxPool2d layers, the numbers in the brackets are the kernel size, stride, and padding. $b$ is the bag size (number of patches).}
    \vspace{0.7mm}
    \label{tab:s2p_small_arch}
    \centering
    \begin{tabular}{llll}
        \toprule
        Layer & Type & Input & Output \\
        \midrule
        1 & Conv2d(4, 1, 0) + ReLu & $b$ x 3 x 28 x 28 & $b$ x 36 x 25 x 25 \\
        2 & MaxPool2d(2, 2, 0) & $b$ x 36 x 25 x 25 & $b$ x 36 x 12 x 12 \\
        3 & Conv2d(3, 1, 0) + ReLu & $b$ x 36 x 12 x 12 & $b$ x 48 x 10 x 10 \\
        4 & MaxPool2d(2, 2, 0) & $b$ x 48 x 10 x 10 & $b$ x 48 x 5 x 5 \\
        - & Flatten & $b$ x 48 x 5 x 5 & $b$ x 1200 \\
        5 & FC + ReLU + Dropout & $b$ x 1200 & $b$ x 512 \\
        6 & FC + ReLU + Dropout & $b$ x 512 & $b$ x 128 \\
        7 & FC + ReLU + Dropout & $b$ x 128 & $b$ x 64 \\
        8 & FC + ReLU + Dropout & $b$ x 64 & $b$ x 7 \\
        - & MIL Mean Aggregation & $b$ x 7 & 7 \\
        \bottomrule
    \end{tabular}
\end{table}
\vspace{-0.2cm}

\vfill

\vspace{-0.2cm}
\begin{table}[!htb]
	\centering
    \caption{S2P Small Architecture; patch size 56. For the Conv2d and MaxPool2d layers, the numbers in the brackets are the kernel size, stride, and padding. $b$ is the bag size (number of patches).}
    \vspace{0.7mm}
    \label{tab:s2p_medium_arch}
    \centering
    \begin{tabular}{llll}
        \toprule
        Layer & Type & Input & Output \\
        \midrule
        1 & Conv2d(4, 1, 0) + ReLu & $b$ x 3 x 56 x 56 & $b$ x 36 x 53 x 53 \\
        2 & MaxPool2d(2, 2, 0) & $b$ x 36 x 53 x 53 & $b$ x 36 x 26 x 26 \\
        3 & Conv2d(3, 1, 0) + ReLu & $b$ x 36 x 26 x 26 & $b$ x 48 x 24 x 24 \\
        4 & MaxPool2d(2, 2, 0) & $b$ x 48 x 24 x 24 & $b$ x 48 x 12 x 12 \\
        - & Flatten & $b$ x 48 x 12 x 12 & $b$ x 6912 \\
        5 & FC + ReLU + Dropout & $b$ x 6912 & $b$ x 512 \\
        6 & FC + ReLU + Dropout & $b$ x 512 & $b$ x 128 \\
        7 & FC + ReLU + Dropout & $b$ x 128 & $b$ x 64 \\
        8 & FC + ReLU + Dropout & $b$ x 64 & $b$ x 7 \\
        - & MIL Mean Aggregation & $b$ x 7 & 7 \\
        \bottomrule
    \end{tabular}
\end{table}
\vspace{-0.2cm}

\vfill

\clearpage

\vspace{-0.2cm}
\begin{table}[!htb]
	\centering
    \caption{S2P Small Architecture; patch size 102. For the Conv2d and MaxPool2d layers, the numbers in the brackets are the kernel size, stride, and padding. $b$ is the bag size (number of patches).}
    \vspace{0.7mm}
    \label{tab:s2p_large_arch}
    \centering
    \begin{tabular}{llll}
        \toprule
        Layer & Type & Input & Output \\
        \midrule
        1 & Conv2d(4, 1, 0) + ReLu & $b$ x 3 x 102 x 102 & $b$ x 36 x 99 x 99 \\
        2 & MaxPool2d(2, 2, 0) & $b$ x 36 x 99 x 99 & $b$ x 36 x 49 x 49 \\
        3 & Conv2d(3, 1, 0) + ReLu & $b$ x 36 x 49 x 49 & $b$ x 48 x 47 x 47 \\
        4 & MaxPool2d(2, 2, 0) & $b$ x 48 x 47 x 47 & $b$ x 48 x 23 x 23 \\
        5 & Conv2d(3, 1, 0) + ReLu & $b$ x 48 x 23 x 23 & $b$ x 56 x 21 x 21 \\
        6 & MaxPool2d(2, 2, 0) & $b$ x 56 x 21 x 21 & $b$ x 56 x 10 x 10 \\
        - & Flatten & $b$ x 56 x 10 x 10 & $b$ x 5600 \\
        7 & FC + ReLU + Dropout & $b$ x 5600 & $b$ x 512 \\
        8 & FC + ReLU + Dropout & $b$ x 512 & $b$ x 128 \\
        9 & FC + ReLU + Dropout & $b$ x 128 & $b$ x 64 \\
        10 & FC + ReLU + Dropout & $b$ x 64 & $b$ x 7 \\
        - & MIL Mean Aggregation & $b$ x 7 & 7 \\
        \bottomrule
    \end{tabular}
\end{table}
\vspace{-0.2cm}

\vfill

\subsection{Training Procedure}
\label{app:training}

All of our models were trained to minimise scene-level root mean square error (RMSE) using the Adam optimiser. The hyperparamater details for learning rate, weight decay, and dropout are given in Table \ref{tab:hyperparams}. We utilised early stopping based on validation performance \textemdash{} if the validation RMSE had not decreased for 5 epochs then we terminated the training procedure and reset the model to the point at which it caused the last decrease in validation loss. Otherwise, training terminated after 30 epochs.

\begin{table}[htb!]
	\centering
    \caption{Model training hyperparameters.}
    \label{tab:hyperparams}
    \centering
    \begin{tabular}{llll}
            \toprule
			Configuration & Learning Rate & Weight Decay & Dropout \\
			\midrule
			ResNet18 & 0.05 & 0.1 & N/A \\
            UNet 224 & \num{5e-4} & \num{1e-5} & 0.25 \\
            UNet 448 & \num{5e-4} & \num{1e-6} & 0.2 \\
            \greyrule
			S2P Small 8 & \num{1e-4} & \num{1e-6} & 0.05 \\
			S2P Medium 8 & \num{1e-4} & \num{1e-5} & 0.35 \\
			S2P Large 8 & \num{1e-4} & \num{1e-5} & 0.25 \\
           \greyrule
			S2P Small 16 & \num{5e-4} & \num{1e-6} & 0.1 \\
			S2P Medium 16 & \num{1e-4} & \num{1e-6} & 0.05 \\
			S2P Large 16 & \num{1e-4} & \num{1e-5} & 0.35 \\
           \greyrule
			S2P Small 24 & \num{1e-4} & \num{1e-5} & 0.05 \\
			S2P Medium 24 & \num{1e-4} & \num{1e-6} & 0.2 \\
			S2P Large 24 & \num{5e-4} & \num{1e-5} & 0.3 \\
			\bottomrule
    \end{tabular}
\end{table}

\vfill

\clearpage

\section{Additional Interpretability Outputs}
\label{sec:app_interpretability}

\begin{figure}[htb!]
    \centering
    \includegraphics[width=0.9\textwidth]{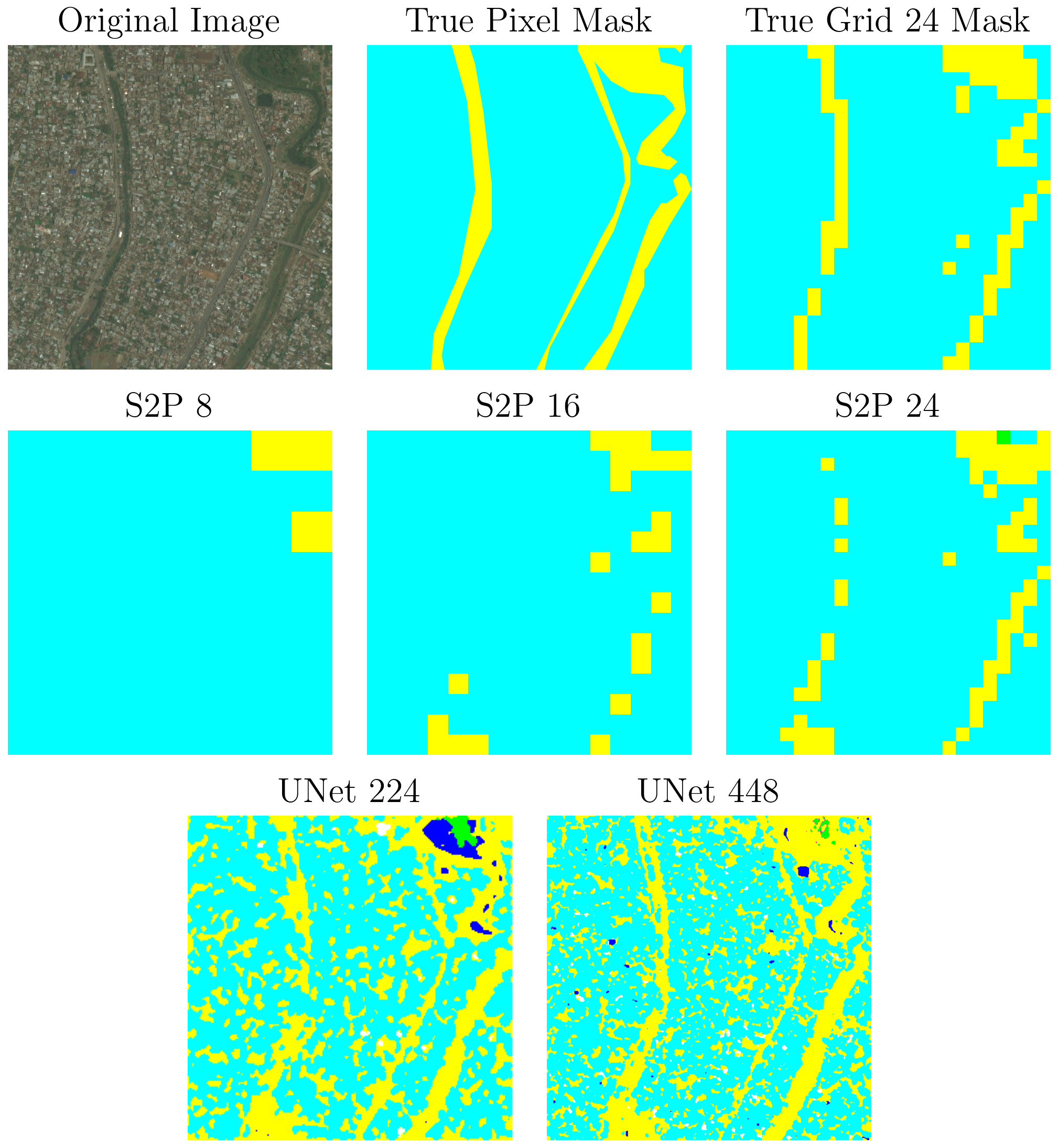}
    \caption{A comparison of patch predictions at different grid sizes. With a grid size of 8, the cells are too large to be able to resolve the fine detail of the agricultural land (yellow) \textemdash{} only the large areas of agricultural land and can be separated from the urban land (blue). Grid sizes of 16 and 24 provide smaller cells, allowing the model to correctly identify more of the agricultural regions. The UNet models are able to resolve the agricultural regions in the true pixel mask, but also pick up on lots of other agricultural regions not labelled in the ground truth. They also make misclassify other areas in the image, identifying water and forest regions. Note this is using unweighted model outputs.}  
    \label{fig:interpretability_comp}
\end{figure}

\begin{figure}[htb!]
    \centering
    \includegraphics[width=\textwidth]{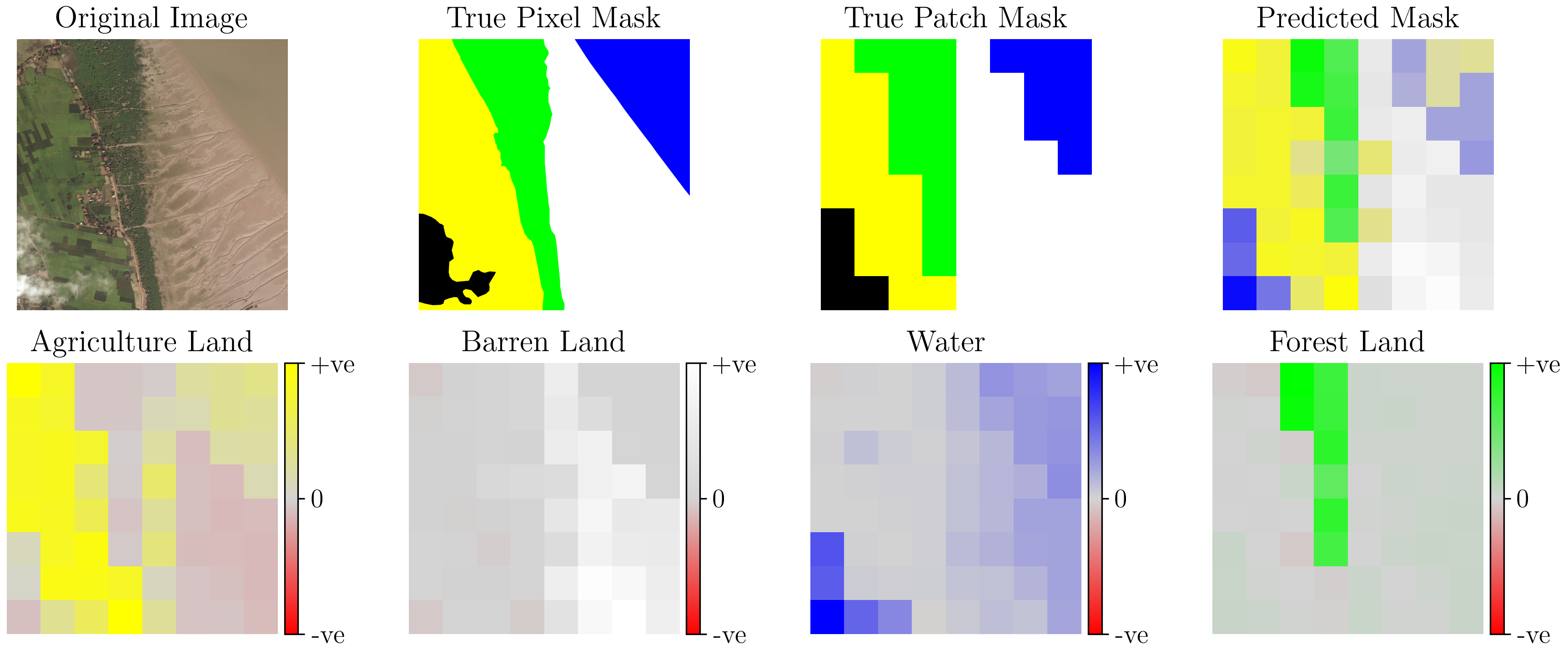}
    \vspace{-0.6cm}
    \caption{An output from an S2P Large 8 model. Cloud covers the bottom left of the original image, so that area has been marked as unknown (black). However, the model labels it water, i.e., it has been unable to correctly learn the unknown class as it occurs very infrequently in this dataset.}  
    \label{fig:interpretability_extra_8}
\end{figure}

\begin{figure}[htb!]
    \centering
    \includegraphics[width=\textwidth]{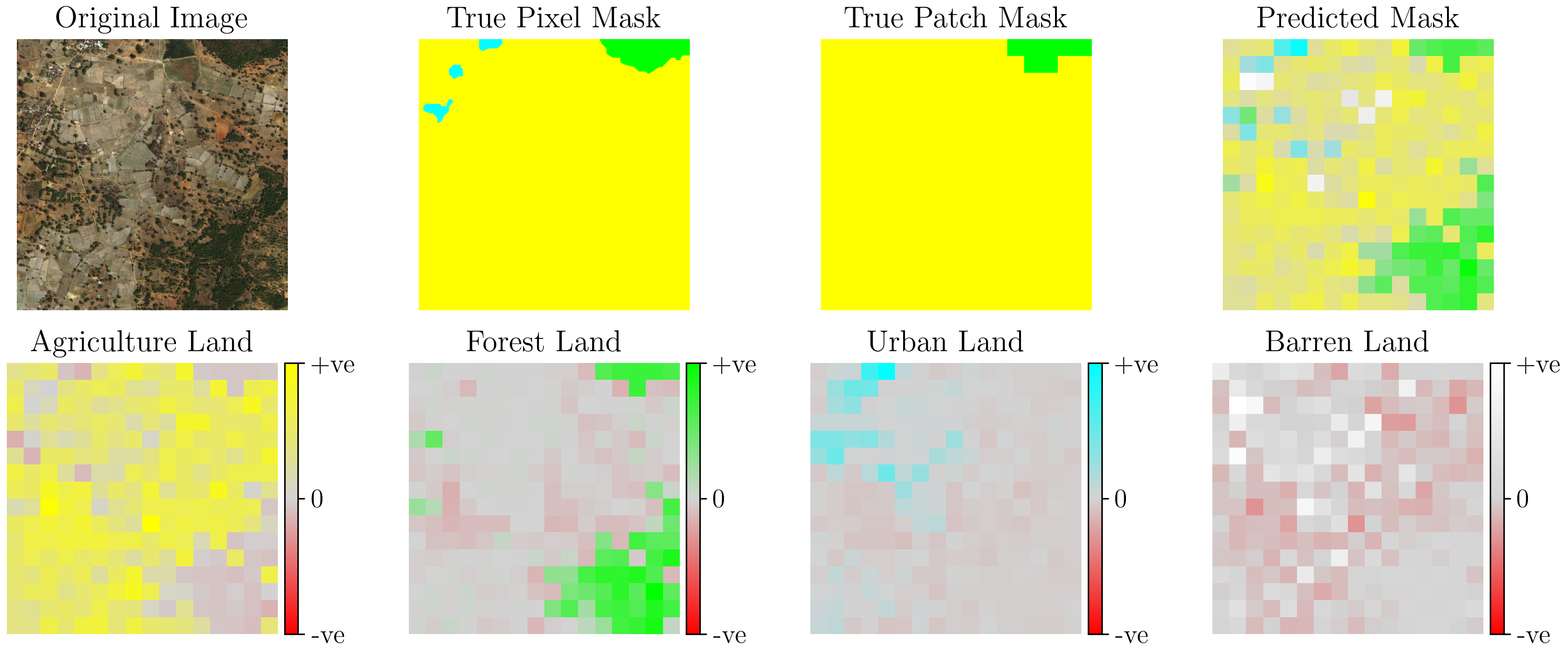}
    \vspace{-0.6cm}
    \caption{An output from an S2P Medium 16 model. This gives an example of potential mislabelling in the ground truth, where the trees in the lower right have not been labelled, but the ones in the upper right have. However, this could be intentional as the two regions have different tree densities. The model predicts both regions as forest, but will be penalised for doing so. Also note that the urban areas disappear in the true patch mask as they are not the majority class in their respective patches. However, the model still predicts those regions as urban, matching the pixel mask but not the patch mask.}  
    \label{fig:interpretability_extra_16}
\end{figure}

\begin{figure}[htb!]
    \centering
    \includegraphics[width=\textwidth]{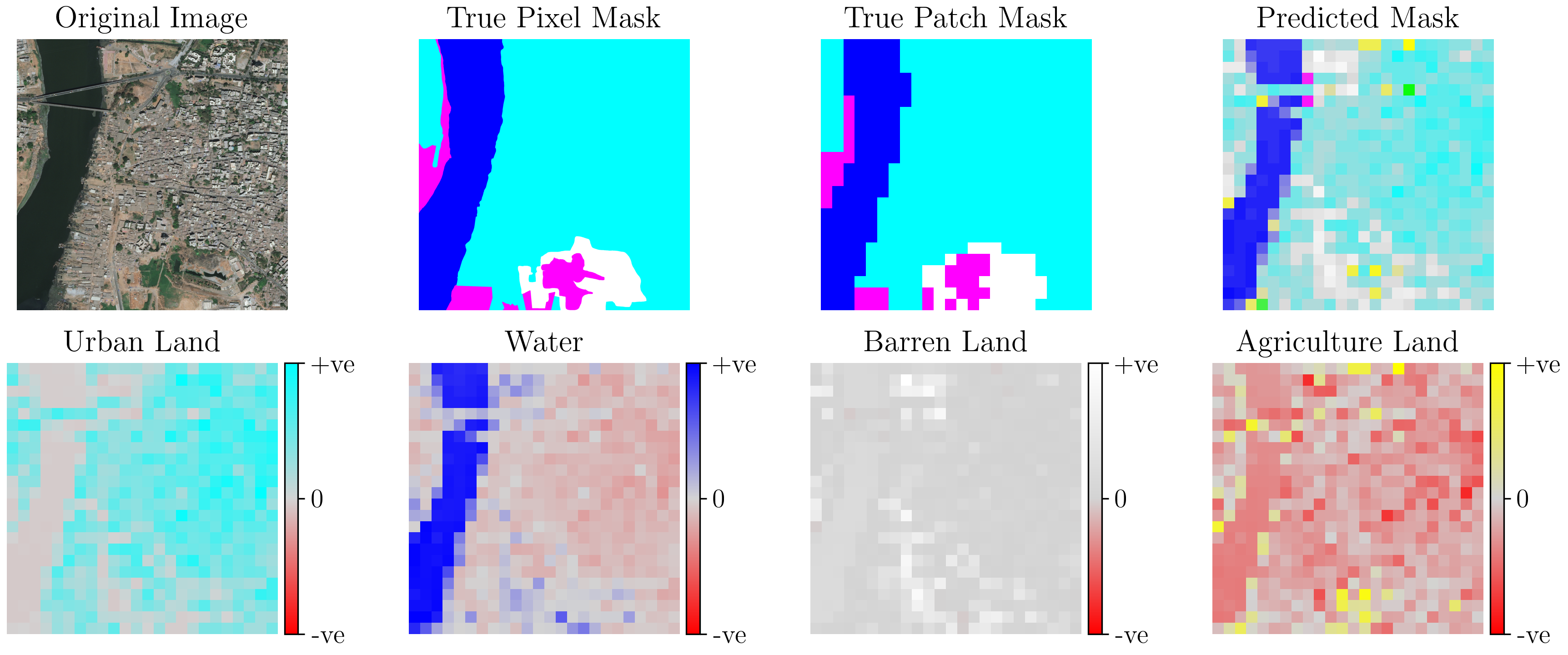}
    \vspace{-0.6cm}
    \caption{An output from an S2P Medium 24 model. Due to the large grid size, the model is able to segment the bridge crossing the water and label it as urban (blue). However, the ground truth labelling omits this, meaning segmenting the bridge in this way is incorrect (according to the original segmentation labels).}  
    \label{fig:interpretability_extra_24}
\end{figure}

\begin{figure}[htb!]
    \centering
    \includegraphics[width=\textwidth]{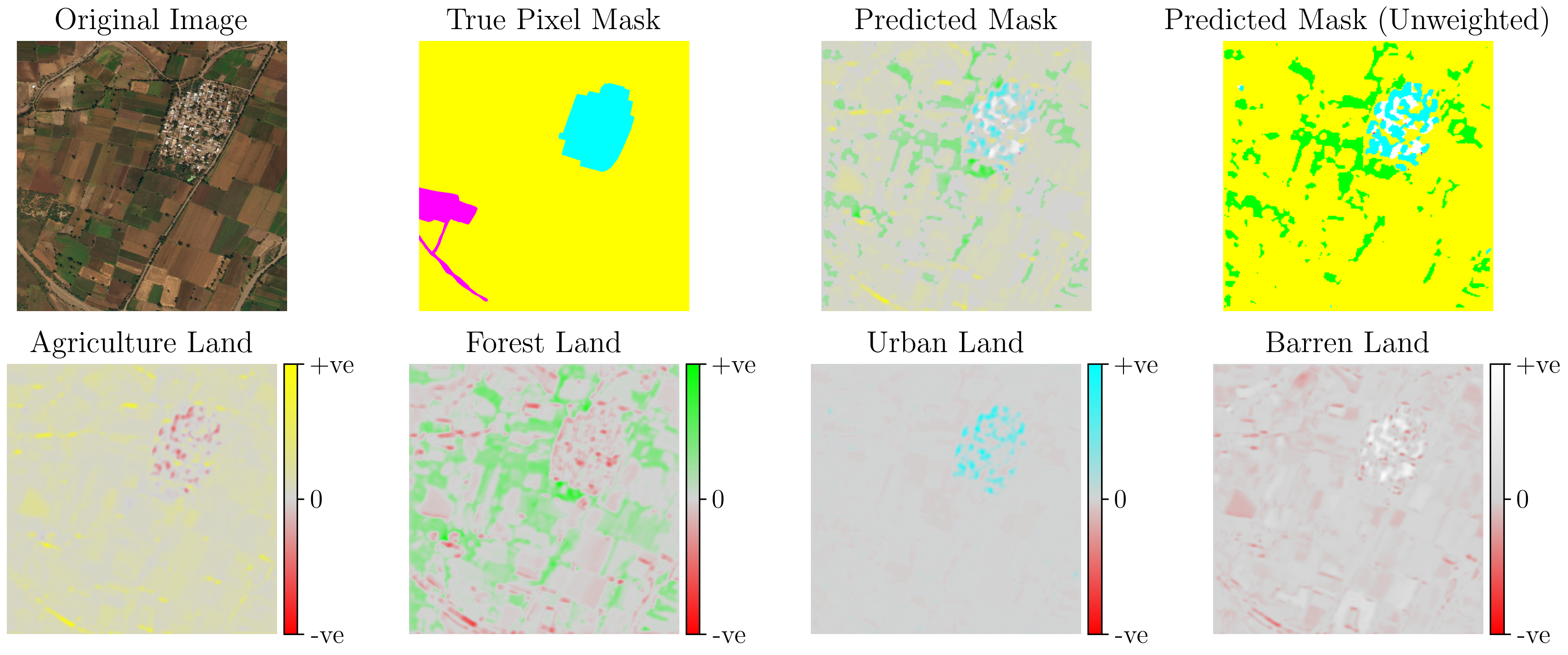}
    \vspace{-0.6cm}
    \caption{An output from a UNet 224 model. The model is able to separate features such as individual buildings in the urban area, and small clusters of trees (that weren't labelled in the true pixel mask). However, it misclassifies the rangeland regions in the bottom left of the image. As the model often has relatively low weighted predictions for some areas, we also provide the unweighted prediction mask.}  
    \label{fig:interpretability_extra_unet224}
\end{figure}

\begin{figure}[htb!]
    \centering
    \includegraphics[width=\textwidth]{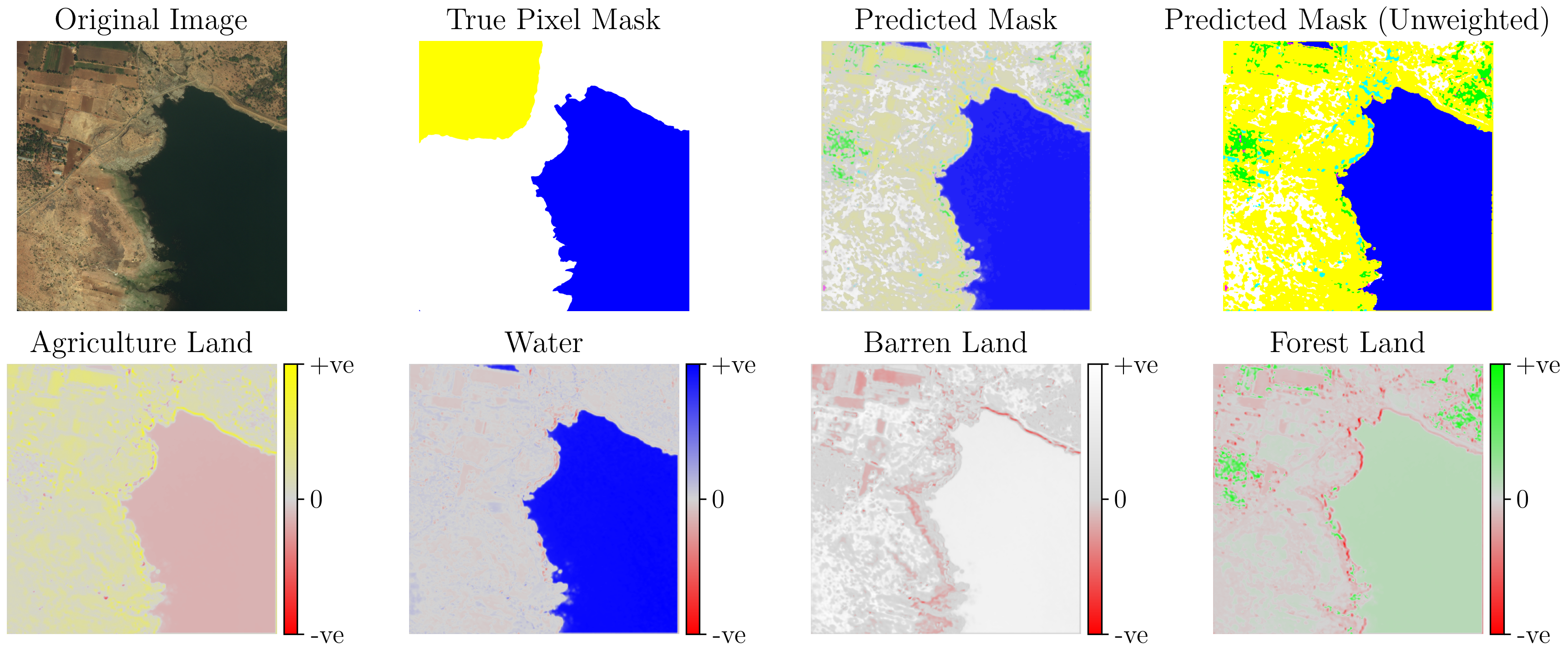}
    \vspace{-0.6cm}
    \caption{An output from a UNet 448 model. The model has been able to segment the area of water with high fidelity, but produces different segmented regions to the true pixel mask for the land. Note on the bottom row, the model also predicts the lower right region as barren and forest, but not as strongly as it (correctly) predicts it as water. This highlights potential model confusion, with smooth blue/green/grey areas leading to misclassification. However, for agricultural land, the model correctly labels the lower right as negative (red), i.e., refuting the prediction of agricultural land for the area of water.}  
    \label{fig:interpretability_extra_unet448}
\end{figure}

Below we provide further outputs of our S2P models. In Figure \ref{fig:interpretability_comp}, we compare the predictions for the different S2P and UNet models. In Figures \ref{fig:interpretability_extra_8} - \ref{fig:interpretability_extra_unet448} we examine edge cases involving potential mislabelling and confusion in model prediction.

\vfill

\vfill

\clearpage

\vfill

\vfill

\vfill

\clearpage

\vfill

\vfill

\vfill

\clearpage

\vfill

\clearpage

\end{document}